%% file: gradient_template.tex
\documentclass[]{gradient} 

\usepackage{array} 

\usepackage{microtype}
\usepackage{makecell}
\usepackage{textgreek} 
\usepackage{graphicx}

\usepackage{booktabs} 
\usepackage[table,xcdraw]{xcolor}
\usepackage{xurl}

\usepackage{hyperref}

\usepackage{subcaption}

\usepackage{tcolorbox}
\tcbuselibrary{listings}

\usepackage{amsmath}
\usepackage{amssymb}
\usepackage{mathtools}
\usepackage{amsthm}

\usepackage{natbib}

\usepackage{colortbl}   
\usepackage{wrapfig}

\usepackage{mathpazo}

\usepackage[linesnumbered,ruled,vlined]{algorithm2e}
\usepackage{listings}

\title{DynaWeb: Model-Based Reinforcement Learning of Web Agents}
\author{
    Hang Ding\textsuperscript{1}
    \quad Peidong Liu\textsuperscript{2} 
    \quad Junqiao Wang\textsuperscript{1} 
    \quad Ziwei Ji\textsuperscript{3} 
    \quad Meng Cao\textsuperscript{4\, 8} 
    \quad Rongzhao Zhang\textsuperscript{5} \\
    \quad Lynn Ai\textsuperscript{6} 
    \quad Eric Yang\textsuperscript{6}
    \quad Tianyu Shi\textsuperscript{6}
    \quad Lei Yu\textsuperscript{7}

}
 
\affiliation{
    \textsuperscript{1}Shanghai Jiao Tong University \quad
    \textsuperscript{2}Sichuan University \quad
    \textsuperscript{3}Hong Kong University of Science and Technology \quad \\
    \textsuperscript{4}McGill University \quad 
    \textsuperscript{5}Shanghai AI Lab \quad
    \textsuperscript{6}Gradient \quad \\
    \textsuperscript{7}University of Toronto \quad
    \textsuperscript{8}Mila - Quebec AI Institute \quad
}

\date{Jan 30, 2026}
\correspondence{jadeleiyu@cs.toronto.edu, ty.shi@mail.utoronto.ca}
\project{ty.shi@mail.utoronto.ca}
\sourcecode{Coming-Soon}

\begin{document}

\abstract{The development of autonomous web agents, powered by Large Language Models (LLMs) and reinforcement learning (RL), represents a significant step towards general-purpose AI assistants. However, training these agents is severely hampered by the challenges of interacting with the live internet, which is inefficient, costly, and fraught with risks. Model-based reinforcement learning (MBRL) offers a promising solution by learning a \textit{world model} of the environment to enable simulated interaction. This paper introduces DynaWeb, a novel MBRL framework that trains web agents through interacting with a web world model trained to predict naturalistic web page representations given agent actions. This model serves as a synthetic web environment where an agent policy can \textit{dream} by generating vast quantities of rollout action trajectories for efficient online reinforcement learning. Beyond free policy rollouts, DynaWeb incorporates real expert trajectories from training data, which are randomly interleaved with on-policy rollouts during training to improve stability and sample efficiency. Experiments conducted on the challenging WebArena and WebVoyager benchmarks demonstrate that DynaWeb consistently and significantly improves the performance of state-of-the-art open-source web agent models. Our findings establish the viability of training web agents through imagination, offering a scalable and efficient way to scale up online agentic RL.}
\maketitle

\input{tex/1_intro}
\input{tex/2_related_work}

\input{tex/3_method}

\input{tex/4_experiments}

\input{tex/5_analysis}

\input{tex/6_conclusion}

\bibliography{gradient}

\appendix
\onecolumn
\input{tex/appendix}

\end{document}

%% file: tex/1_intro.tex
\section{Introduction}
The paradigm of artificial intelligence is rapidly shifting toward proactive, agentic systems that can autonomously execute complex, long-horizon tasks in open-ended environments.
Large Language Models (LLMs) have emerged as a powerful backbone for such agents, enabling rich reasoning, flexible action generation, and natural language interaction.
In the web domain, LLM-based agents have demonstrated strong capabilities in navigating real websites and accomplishing user-specified goals through multi-step interaction, fueled by advances in prompting, structured reasoning, and action abstractions~\citep{yaoreact, webarena, webvoyager}.
Beyond imitation learning, recent work has shown that training web agents with online reinforcement learning (RL) can substantially improve robustness, exploration, and long-horizon decision-making, allowing agents to surpass static demonstrations and adapt through trial-and-error interaction~\citep{wei2025webagentr1trainingwebagents, qi2025webrltrainingllmweb, zhuang2025workforceagentr1incentivizingreasoningcapability}.

Despite its promise, the effectiveness of online RL for web agents is fundamentally constrained by the cost and risk of real-environment interaction.
Collecting large-scale, on-policy experience requires agents to interact directly with the live internet, which is inefficient, expensive, and difficult to control at scale.
\begin{wrapfigure}{l}{0.6\columnwidth}
  \vspace{-0.8em}
  \centering
  \includegraphics[width=0.6\columnwidth]{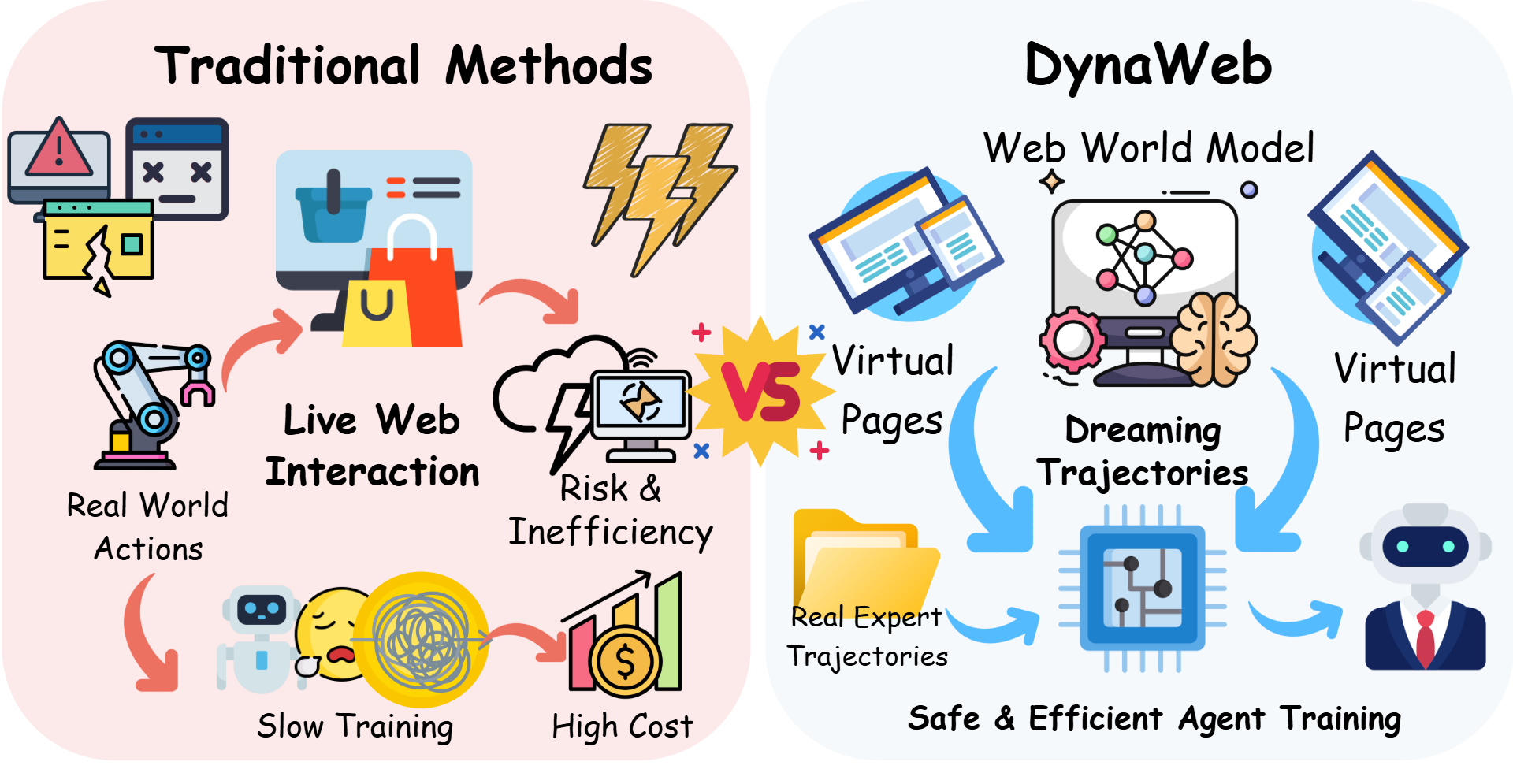}
  \vspace{-0.6em}
  \caption{
  Comparison between traditional web agent training via live web interaction and DynaWeb.
  By replacing risky and inefficient real-world interaction with a learned web world model,
  DynaWeb enables imagination-driven training using virtual pages and dreamed trajectories,
  optionally augmented with real expert data, resulting in safer and more efficient agent optimization.
  }
  \vspace{-0.8em}
  \label{fig:teaser}
\end{wrapfigure}
During training, agents may trigger irreversible actions such as unintended purchases, account modifications, or data submissions, while also facing non-deterministic page dynamics, transient failures, and external interference.
These challenges severely limit the practicality of pure online RL, rendering large-scale policy optimization both costly and hazardous in real-world web environments~\citep{webarena,qi2025webrltrainingllmweb}.
As a result, a central open question emerges: how can we retain the benefits of online reinforcement learning for web agents while dramatically reducing reliance on direct interaction with the live web?
A natural direction is to replace expensive and risky real-environment interaction with a learned, controllable surrogate that can faithfully approximate web dynamics.
To this end, recent work has begun to explore web \emph{world models}—learned simulators of web environments.
So far, however, their role has been largely auxiliary.
Some approaches employ world models at inference time for short-horizon look-ahead and action evaluation, treating them as decision-time reasoning tools rather than learning substrates~\citep{webdreamer,wei2025webagentr1trainingwebagents}.
Others use world models to synthesize offline trajectories for supervised fine-tuning or imitation-style training, decoupling model-generated experience from on-policy optimization~\citep{fang2025webevolver,Explorer}.
What remains missing is an \emph{online} model-based reinforcement learning (MBRL) paradigm for web agents, in which imagined rollouts serve as first-class experience that directly augments on-policy policy optimization.

In this work, we revisit classical model-based reinforcement learning through the lens of modern web agents.
Inspired by the Dyna architecture~\citep{sutton1991dyna} and imagination-based learning frameworks such as Dreamer~\citep{hafner2020dreamcontrollearningbehaviors}, we propose \textbf{DynaWeb}, an MBRL framework that elevates a web world model from a planning or data-generation tool to a core component of online reinforcement learning.
DynaWeb treats the world model as a controllable synthetic web environment that can replace or augment costly real interaction.
Concretely, an LLM-based web world model functions as a learned web server: conditioned on the current page representation and an agent action, it predicts realistic next-state page representations and provides task-level feedback signals for policy optimization.
By training web agents on a mixture of real and imagined experience, DynaWeb enables scalable, on-policy reinforcement learning through imagination while preserving the benefits of interactive learning.

Crucially, DynaWeb combines two complementary sources of training experience.
In addition to policy-driven imagined rollouts generated by the web world model, we directly incorporate fully real expert trajectories sampled from existing training data.
These expert trajectories are entirely independent of the world model and correspond to ground-truth web interactions.
During training, real expert trajectories are randomly interleaved with imagined rollouts and real on-policy interaction, allowing the agent to benefit from high-quality demonstrations while substantially reducing reliance on costly live web exploration.
This simple but effective interleaving strategy preserves the on-policy learning signal and enables efficient online reinforcement learning with significantly fewer real-environment interactions.

Our contributions are summarized as follows:
\begin{itemize}
  \item We train a web world model that predicts naturalistic web page state transitions in the form of structured accessibility tree representations, enabling realistic simulation of web environment dynamics without live web interaction.
  \item We show that policy-driven imagined rollouts generated by a learned web world model can be directly used as on-policy training experience for reinforcement learning of web agents.
  \item We propose \textbf{DynaWeb}, a model-based reinforcement learning framework that integrates fully real expert trajectories from training data with imagined rollouts from a world model, enabling effective policy optimization without interacting with the live web.
  \item Through extensive experiments and analysis, we demonstrate consistent performance improvements on WebArena and WebVoyager, and provide empirical insights into the design and use of world-model-based imagination for training web agents.
\end{itemize}

%% file: tex/2_related_work.tex
\section{Related work}

\paragraph{Web Agent}
Recent advances in web agents are largely driven by (multimodal) large language models (LLMs) serving as the core decision-making backbone \citep{llama3, leopard, gpt4, claude37,10800533,zeng-etal-2025-bridging,codes2026,li2025lion,li2025cogvla,li2025semanticvla,ding2026arkanswercentricretrievertuning}. On top of these models, reasoning and interaction frameworks such as ReAct~\cite{yaoreact}, MCP~\citep{MCP}, and Cognitive Kernel~\citep{cognitive_kernel} enable structured multi-step web actions. Web agents are commonly evaluated on interactive benchmarks including WebShop~\citep{webshop}, Mind2Web~\citep{mind2web}, WebArena~\citep{webarena}, VisualWebArena~\citep{visualwebarena}, WebVoyager~\citep{webvoyager}, WebWalker~\citep{WebWalker}, and MMInA~\citep{MMInA}.

Beyond off-the-shelf prompting, a broad line of work improves web agents via data scaling and stronger agent training pipelines. Data-centric efforts such as Explorer~\citep{Explorer}, NNetNav~\citep{NNetNav}, and InSTA~\citep{DBLP:journals/corr/abs-2502-06776} collect or synthesize high-quality interaction data, while recent agent foundation models and deep research agents push toward more general web interaction at scale, e.g., WebThinker~\citep{li2025webthinker}, WebDancer~\citep{wu2025webdancer}, WebSailor~\citep{li2025websailor}, WebShaper~\citep{tao2025webshaper}, Cognitive Kernel-Pro~\citep{cognitivekernel-pro}, MiroFlow~\citep{MiroFlow2025}, and multimodal research agents such as WebWatcher~\citep{geng2025webwatcherbreakingnewfrontier}. In parallel, inference-time optimization (e.g., tree search and self-reflection) further improves decision-making without additional training~\citep{AgentTreeSearch, AgentQ, Exact, LATS, webpilot, Reflexion,zeng2025lensllmunveilingfinetuningdynamics,lin2025planbudgeteffectiveefficient,zeng2026halluguarddemystifyingdatadrivenreasoningdriven,zhou2024adversarial,zhouboosting,liu2026discoragdiscourseawareretrievalaugmentedgeneration}.

\paragraph{World Models}
World models originate from model-based reinforcement learning \citep{world_model} and have recently been revisited as a powerful abstraction for agent reasoning and planning \citep{DBLP:journals/corr/abs-2408-14837, DBLP:conf/nips/AlonsoJMKSPF24, DBLP:journals/corr/abs-2305-14223,zhou2022understanding,zhou2025class,zhou2026dynamic,yu2024robust,yu2024mechanistic,li2026sesearchselfevolvingsearchagent}. With the emergence of large language models (LLMs), several works treat LLMs as implicit world models capable of simulating future states. For general reasoning, RAP~\citep{RAP_world_model} integrates LLM-based world modeling with Monte Carlo Tree Search to explore hypothetical trajectories, while WKM~\citep{agent_world_model} distills structured world knowledge from agent trajectories to guide planning.

In web environments, this paradigm has been adapted by methods such as WebDreamer~\citep{webdreamer} and WMA~\citep{web_agent_world_model}, which use LLMs to predict the outcomes of web actions via natural language simulation. However, these approaches primarily employ world models at inference time, functioning as planning or prompting modules to assist action selection. Even when multi-step simulation is performed, the imagined rollouts are used only to guide immediate decisions rather than to improve the agent policy itself through training.

In contrast, our work positions the web world model as a core component of the learning process. By training a dedicated world model and using it to generate multi-step imagined trajectories for policy optimization, DynaWeb moves beyond inference-time reasoning and enables genuine model-based reinforcement learning for web agents. This design directly addresses the data inefficiency and cost of real-environment interaction highlighted in prior RL-based web agents, and aligns with our goal of scalable, imagination-driven agent training.

\paragraph{Reinforcement Learning of Agents}
Reinforcement learning (RL) provides a direct mechanism to improve agent policies from interactive feedback, which is particularly appealing for web environments with long-horizon, multi-step decision making \citep{wei2025webagentr1trainingwebagents, qi2025webrltrainingllmweb, wu2025webdancer, zhuang2025workforceagentr1incentivizingreasoningcapability,cao2024enhancing,tao2024rolecraft,ji2025calibrating,hong2025reasoning,ding-etal-2026-scirag}. A representative end-to-end online RL approach is WebAgent-R1~\citep{wei2025webagentr1trainingwebagents}, which optimizes multi-turn web interaction policies using outcome-based rewards and scalable trajectory sampling (e.g., multi-group GRPO~\citep{shao2024deepseekmathpushinglimitsmathematical}). Complementary to direct end-to-end optimization, WebRL~\citep{qi2025webrltrainingllmweb} emphasizes self-evolving curriculum design and result-supervised feedback to continually generate training tasks and improve agent robustness. 

Other methods adapt RL objectives or combine RL with additional supervision to better shape reasoning and planning behaviors. WebDancer~\citep{wu2025webdancer} leverages structured supervision from QA-style signals and policy optimization (e.g., DAPO~\citep{yu2025dapoopensourcellmreinforcement}) to internalize actionable reasoning traces. WorkForceAgent-R1~\citep{zhuang2025workforceagentr1incentivizingreasoningcapability} integrates behavior cloning with GRPO-style optimization to jointly enhance single-step reasoning and multi-step planning. Beyond pure online RL, joint training pipelines that combine SFT and RL have also shown effectiveness. For instance, AutoWebGLM~\citep{10.1145/3637528.3671620} integrates SFT, RL, and rejection sampling fine-tuning, with curriculum learning to bootstrap basic web navigation skills before subsequent RL refinement\citep{shayegani2025misaligned,qiu2025hallucination,xu2025sedm,ding2025rolermbenchrolermreward}.

Despite their strong results, these RL-based agents typically rely on large amounts of real-environment interaction, which can be expensive, unstable, and risky at scale---a key motivation for our imagination-driven training framework.

%% file: tex/3_method.tex
\begin{figure*}[t]
  \centering
  \includegraphics[width=\textwidth]{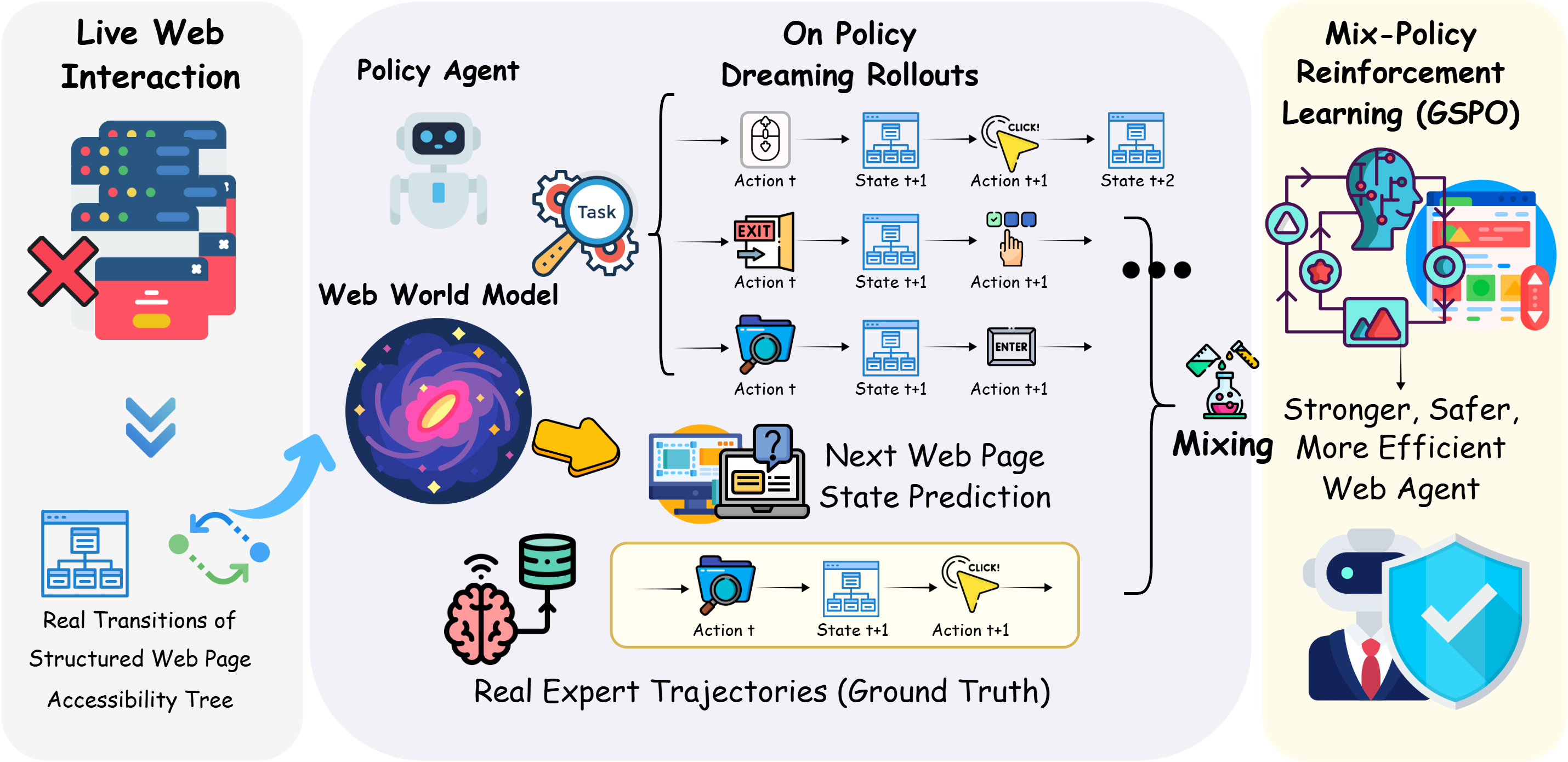}
  \caption{
  \textbf{Overview of DynaWeb.}
  DynaWeb trains web agents via imagination-driven, model-based reinforcement learning.
  A learned web world model serves as a synthetic environment, enabling the agent to generate multi-step imagined rollouts without interacting with the live web.
  These imagined trajectories are mixed with a small fraction of real expert trajectories to stabilize learning.
  The agent policy is optimized using sequence-level policy optimization, allowing efficient and robust credit assignment for long-horizon web tasks with sparse terminal rewards.
  }
  \label{fig:dynaweb_overview}
\end{figure*}

\section{Method}\label{sec:method}

\subsection{Problem Formulation}

We formulate the web agent task as a Partially Observable Markov Decision Process (POMDP)
$(\mathcal{S}, \mathcal{A}, \mathcal{O}, \mathcal{T}, \mathcal{R})$.
Given a natural language query $q$ and a target website $w$, the agent is required to complete a multi-step web interaction to satisfy the task objective.
The state space $\mathcal{S}$ represents the full underlying web environment, which is not directly observable by the agent.
Instead, at each time step $t$, the agent receives a partial observation $o_t \in \mathcal{O}$ obtained via an observation function
$o_t = \Omega(s_t)$, where $\Omega(\cdot)$ extracts visible information such as the current URL and web elements from state $s_t$.

The action space $\mathcal{A}$ consists of atomic browser operations, including \texttt{click}, \texttt{type}, \texttt{goback}, \texttt{scroll up/down}, and \texttt{stop}.
Executing an action $a_t \in \mathcal{A}$ deterministically advances the environment according to the transition function $\mathcal{T}$:
\begin{equation}
s_{t+1} = \mathcal{T}(s_t, a_t), \quad o_{t+1} = \Omega(s_{t+1}).
\end{equation}

Following prior work~\citep{cognitive_kernel}, we represent each observation $o_t$ using an accessibility tree that captures the structured layout of visible web elements.
The agent is parameterized by a large language model (LLM) policy $\pi_\theta$, which conditions on the task query $q$, the interaction history, and system instructions $I$ to generate reasoning traces and actions.
At time step $t$, the agent samples a chain-of-thought $h_t$ together with an action $a_t$ as
\begin{equation}
(h_t, a_t) \sim \pi_\theta(\cdot \mid I, q, o_{1:t}, h_{1:t-1}, a_{1:t-1}).
\end{equation}

An interaction episode produces a trajectory
$\tau = (o_1, h_1, a_1, \ldots, o_T, h_T, a_T)$,
where $T$ denotes the total number of interaction steps.
Upon termination, the trajectory is assigned a scalar reward $\hat{r}(\tau, q) \in [0,1]$ based on task completion, obtained via self-assessment or environment-specific evaluation.

\subsection{Web World Model}

We introduce a web world model that approximates the transition dynamics of web environments by predicting how visible web states evolve in response to agent actions. 
Given the partial observation $o_t$ represented as an accessibility tree and an action $a_t \in \mathcal{A}$, the world model predicts the next observation $\hat{o}_{t+1}$:
\begin{equation}
\hat{o}_{t+1} \sim p_\phi(\cdot \mid o_t, a_t, I),
\end{equation}
where $p_\phi$ is a large language model parameterized by $\phi$.
Rather than learning latent dynamics, the model operates directly in the observation space, generating naturalistic web page representations that can be directly consumed by the agent policy during imagined rollouts.

As shown in previous studies, simply using textual observations to represent web environment states and use them as training objectives would be problematic, as state transitions in websites often involve modifying only a part of the previous observation, so $o_{t+1}$ often remains highly similar to $o_{t}$, and predicting the entire next textual observation may result in low information gain. We therefore decompose the task of next web state prediction into two subtasks: first, we ask the world model to predict a set of free-from natural language description of \textit{state changes} $\Delta(o_{t}, o_{t+1})$ resulted by the action $a_t$, and then apply $\Delta(o_{t}, o_{t+1})$ on the current state $o_t$ to get $o_{t+1}$. We could therefore only train the world model on the first reasoning subtask of predicting state change descriptions, and leverage its instruction following capability to perform the second subtask of state altering during inference. 

The world model is trained using real web interaction trajectories collected from the StanfordNLP/NNetNav \citep{NNetNav} dataset.
We apply a data cleaning pipeline to filter out erroneous or incomplete trajectories, including missing observations, invalid actions, or inconsistent state transitions.
We then prompt GPT-oss-120b with each valid transition $(I, o_t, a_t, o_{t+1})$ to obtain state change descriptions $\Delta(o_{t}, o_{t+1})$ as well as the model reasoning trace $r$ before generating $\Delta$. The world model LLM is trained as a reasoning model to predict the ground-truth reasoning trace as well as the subsequent state changes, conditioned on the current accessibility tree and the executed action: 
\begin{align}
    \mathcal{L}_{\phi} =  \sum\limits_{(I, o_t, a_t, r, \Delta)} - \log p_{\phi}(r, \Delta | I, o_t, a_t)
\end{align}
This learned world model can therefore serve as a reusable simulator for generating multi-step imagined trajectories, which we leverage for model-based reinforcement learning in the following section.

\subsection{DynaWeb: Model-based RL of Web Agents}

DynaWeb trains web agents through model-based reinforcement learning by relying on imagined rollouts generated by a learned web world model, eliminating the need for live web interaction during training.
Given a task query $q$ and an initial observation $o_1$, the agent policy $\pi_\theta$ interacts with the world model as a synthetic environment to produce multi-step trajectories.
At each step $t$, the agent samples an action
\begin{equation}
a_t \sim \pi_\theta(\cdot \mid o_{1:t}, h_{1:t-1}, a_{1:t-1}, q),
\end{equation}
and the world model predicts the next observation
\begin{equation}
\hat{o}_{t+1} \sim p_\phi(\cdot \mid \hat{o}_t, a_t, q), \quad \hat{o}_1 = o_1.
\end{equation}
Iterating this process yields an imagined trajectory
\begin{equation}
\hat{\tau} = (\hat{o}_1, h_1, a_1, \ldots, \hat{o}_T, h_T, a_T),
\end{equation}
without interacting with the live web environment.

To enable reinforcement learning on imagined trajectories, a task-level reward signal is obtained via model-based self-assessment conditioned on the task query.
After trajectory termination, a task-level completion reward $\hat{r}(\hat{\tau}, q) \in \{0,1\}$ is obtained via model-based self-assessment with respect to the task query, indicating whether the task objective has been achieved.

We treat the return as $G(\hat{\tau})=\hat{r}(\hat{\tau}, q)$ and use the resulting imagined rollouts as valid training data for policy gradient optimization, substantially reducing reliance on real-environment interaction while preserving task-level supervision.

Beyond policy-driven imagined rollouts, DynaWeb additionally incorporates fully real expert trajectories sampled from training data to improve training stability and sample efficiency.
These expert trajectories correspond to ground-truth web interactions and are entirely independent of the world model.
In practice, real expert trajectories are randomly interleaved with policy-driven imagined rollouts during training, allowing the agent to benefit from high-quality demonstrations while learning under a simulated environment.

To optimize the agent policy from this mixture of real and imagined experiences, we adopt Group Sequence Policy Optimization (GSPO)~\citep{zheng2025groupsequencepolicyoptimization}.
Concretely, GSPO lifts importance sampling from the token level to the rollout (sequence) level by assigning each trajectory a single sequence-level ratio, which is then applied uniformly to all tokens in the trajectory.
Let $\{\hat{\tau}^{\,i}\}_{i=1}^G$ denote a group of rollouts sampled from $\pi_{\theta_{\text{old}}}$ (either imagined via the world model or real expert trajectories), and let $y^{\,i}$ be the serialized token sequence produced by the policy along $\hat{\tau}^{\,i}$ (i.e., concatenating all generated reasoning traces and actions).
GSPO optimizes the clipped objective
\begin{equation}
\small
\mathcal{J}_{\text{GSPO}}(\theta)
=
\mathbb{E}_{q \sim \mathcal{D},\, \{\hat{\tau}^{\,i}\}_{i=1}^G \sim \pi_{\theta_{\text{old}}}(\cdot \mid q)}
\!\left[
\frac{1}{G}\sum_{i=1}^G
\min\!\Big(
s^{\,i}(\theta)\,\hat{A}^{\,i},\ 
\operatorname{clip}\!\big(s^{\,i}(\theta),\,1-\varepsilon,\,1+\varepsilon\big)\,\hat{A}^{\,i}
\Big)
\right]
\end{equation}

where $\hat{A}^{\,i}$ denotes the (trajectory-level) advantage computed from the terminal return $G(\hat{\tau}^{\,i})$.
The sequence-level ratio $s^{\,i}(\theta)$ is defined as the geometric mean of token-wise likelihood ratios:
\begin{equation}
\small
s^{\,i}(\theta)
=
\Bigg(
\frac{\pi_{\theta}\!\big(y^{\,i}\mid q,o_1\big)}{\pi_{\theta_{\text{old}}}\!\big(y^{\,i}\mid q,o_1\big)}
\Bigg)^{\!\!1/\lvert y^{\,i}\rvert}
=
\exp\!\Bigg(
\frac{1}{\lvert y^{\,i}\rvert}
\sum_{k=1}^{\lvert y^{\,i}\rvert}
\log r_k^{\,i}(\theta)
\Bigg),
\end{equation}
with token-level ratios
\begin{equation}
\small
r_k^{\,i}(\theta)
=
\frac{\pi_{\theta}\!\big(y_k^{\,i}\mid q,o_1, y_{<k}^{\,i}\big)}
{\pi_{\theta_{\text{old}}}\!\big(y_k^{\,i}\mid q,o_1, y_{<k}^{\,i}\big)}.
\end{equation}

%% file: tex/4_experiments.tex
\section{Experiments}

\subsection{Setups}
\paragraph{Benchmarks and metrics}
We evaluate our DynaWeb agents against baselines on two challenging benchmarks of web navigation tasks: 1) WebArena \citep{webarena}, which includes 812 real-life tasks in simulated web environments across five different websites and four application domains; 2) WebVoyager \citep{webvoyager}, a semi-automatically generated dataset comprising 643 open-ended web tasks from 15 commonly accessed live websites. The main evaluation metric, Success Rate (SR), is calculated as the percentage of the user instructions that are successfully accomplished by the generated agent trajectory. For WebArena, to ensure fair comparison and reproducibility, we conduct our experiments using the official environment hosted on an Amazon Web Services (AWS) EC2 instance pre-configured with a Docker environment. For WebVoyager, since some live websites became no longer accessible in our evaluation web environment, either due to geographical locations or IP blocks, we filter out these failed tasks for our experiments. To ensure robustness, we follow \citep{fang2025webevolver} by running our experiments roughly at the same time window twice and report the average results. 
\paragraph{Web Agents}
We use a state-of-the-art open-source NNetNav web agent family by \citep{NNetNav} fine-tuned from a Llama-3.1-8B as the backbone of our DynaWeb agent. NNetNav provides two web agent models trained live and simulated web environments via supervised fine-tuning datasets of web navigation trajectory. For each of our two evaluation tasks, we take the corresponding NNetNav agent as the initialization checkpoint, and train on the DynaWeb model-based RL environment simulated by a WWM. During training, we implement expert-conditioned rollouts by randomly replacing 50\% of dreamed trajectories with gold-standard trajectories sampled from the NNetNav supervised fine-tuning datasets.

\input{tabs/wa}
\paragraph{World Models}
For web world models, we fine-tune GPT-oss-120b using the curated next web state prediction datasets as explained in section \ref{sec:method}, resulting in a live WWM for WebVoyager and a simulated WWM for WebArena. Each WWM is trained via supervised fine-tuning for one epoch using the LlamaFactory library. During GSPO rollouts, to reduce the effect of WWM hallucinations, we generate synthetic multi-step trajectories up to 5 steps, with early termination if the world model generates a terminal state. To further ensure that DynaWeb training covers web world states across various stages of navigation tasks, we generate synthetic trajectories with initial states that are randomly sampled from the trajectory data in the NNetNav-live or NNetNav-wa SFT dataset (i.e., including both initial and intermediate states of gold-standard trajectories).
\paragraph{Baselines}
We evaluated DynaWeb against several representative baselines, described as follows.
1) \textit{Base models:}
We include an open-source base agent built on Llama-3.1-8B-Instruct with vanilla chain-of-thought prompting, which serves as a minimal reasoning-enabled baseline without additional training or planning mechanisms.
2) \textit{Proprietary model:}
We report results from GPT-4o as a strong closed-source baseline, representing the performance of state-of-the-art proprietary web agents under the same evaluation protocol.
\textit{Supervised fine-tuning (SFT):}
We consider supervised fine-tuning on offline successful web navigation trajectories.
This includes the NNetNav agents of \citep{NNetNav}, trained on expert-collected demonstrations,
as well as Go-Browse\citep{gandhi2025gobrowsetrainingwebagents}, where successful trajectories are automatically collected via structured
exploration and used for standard supervised training without reinforcement learning.

4) \textit{Offline reinforcement learning:}
The offline-RL baseline refers to Llama-3.1-8B-based agents trained using the WebRL framework by \citep{qi2025webrltrainingllmweb}. In this setting, the agent first interacts with online web environments to collect exploration trajectories, which are then scored by a trained outcome-based reward model and used for offline reinforcement learning.
5) \textit{Inference-time lookahead (ITL):}
The ITL baseline is an Llama-3.1-8B-based agent equipped with a web world model at inference time. The agent performs inference-time policy optimization by simulating multiple candidate future states at each action step and selecting the action with the highest expected advantage.

\input{tabs/wy}

\subsection{Main Results}
As shown in Table~\ref{tab:webarena-results}, \textbf{DynaWeb} achieves the strongest overall performance on WebArena, improving the average success rate from \underline{26.7} (Offline-RL) to \textbf{31.0}, corresponding to a relative gain of \textbf{16.1\%}. Notably, these gains are broad rather than localized: DynaWeb attains the best results on \texttt{Reddit}, \texttt{Gitlab}, \texttt{CMS}, and \texttt{Shopping}, indicating that its advantages are not tied to a particular site template or interaction pattern, but generalize across heterogeneous interfaces and task objectives. Compared with Offline-RL and ITL, whose strengths tend to manifest under specific exploration regimes, DynaWeb more consistently transforms partial and imperfect exploration signals into effective learning progress. This suggests that integrating imagined experience into on-policy optimization provides a more reliable training signal for complex, multi-step web interactions.

As shown in Table~\ref{tab:webvoyager-results}, this trend largely carries over to the live, open-world setting of WebVoyager. DynaWeb achieves the best performance on \texttt{All Rec} and outperforms competing methods on a wide range of individual websites, including \texttt{Amazon}, \texttt{Apple}, \texttt{BBC News}, \texttt{Cambridge Dict}, \texttt{Coursera}, \texttt{Google Map}, \texttt{Google Search}, \texttt{HugingFace}, and \texttt{Wolfram$\alpha$}. These results demonstrate that the benefits of model-based, imagination-driven training persist even when the environment is dynamic, partially observable, and less controlled. At the same time, the remaining gaps are informative: DynaWeb underperforms the strongest baselines on a few sites such as \texttt{ArXiv} and \texttt{GitHub}, which typically require longer-horizon planning with highly branching actions and rapidly evolving page states. Overall, the results indicate that DynaWeb offers a robust and transferable improvement for web agents, while highlighting long-horizon world modeling and highly dynamic UI handling as the primary challenges for further progress.

%% file: tabs/wa.tex
\begin{table*}[t]
\caption{
Success rate (\%) on self-hosted websites in WebArena.
Results are reported for five representative website domains (Reddit, GitLab, Maps, CMS, and Shopping).
DynaWeb achieves the highest average success rate and consistently outperforms strong supervised, RL, and inference-time lookahead baselines across most domains.
Best results are in bold, and second-best are underlined.
}
\label{tab:webarena-results}
\centering
\setlength{\extrarowheight}{6pt}
\resizebox{0.85\linewidth}{!}{%
\begin{tabular}{
@{}
p{4.2cm}
*{5}{>{\centering\arraybackslash}p{1.6cm}}
>{\centering\arraybackslash}p{1.9cm}
@{}
}
\multicolumn{7}{>{\columncolor{gray!15}}c}{\textbf{\textit{Self-hosted Websites (WebArena)}}} \\
\toprule
\textit{Method}
& \texttt{Reddit}
& \texttt{Gitlab}
& \texttt{Map}
& \texttt{CMS}
& \texttt{Shopping}
& \textit{Average SR} \\
\midrule
Llama3.1-8B-Instruct CoT
& 0.0 & 3.3 & 3.3 & 2.9 & 11.1 & 4.1 \\
GPT-4o
& 10.5 & 10.0 & \textbf{20.0} & 20.0 & 11.1 & 14.3 \\
Qwen2.5-32B
& 10.5 & 20.0 & \underline{19.2} & 19.2 & 17.8 & 17.3 \\
NNetNav
& 16.4 & 8.7 & 11.4 & 16.7 & 25.9 & 15.8 \\
Go-Browse
& 30.7 & 15.3 & 17.9 & 25.3 & 22.4 & 22.3 \\
WebRL
& \underline{35.6} & \underline{23.5} & 19.1 & \underline{30.6} & 24.7 & \underline{26.7} \\
ITL
& 26.5 & 18.0 & 12.6 & 23.6 & \underline{31.3} & 22.4 \\
\midrule
\textbf{DynaWeb-8B (Ours)}
& \textbf{43.8} & \textbf{28.7} & 17.8 & \textbf{31.5} & \textbf{33.2} & \textbf{31.0} \\
\bottomrule
\end{tabular}%
}
\end{table*}

%% file: tabs/wy.tex

\begin{table*}[t]
\caption{
Success rate (\%) on live websites in WebVoyager.
Results are reported across a diverse set of commonly accessed real-world websites.
DynaWeb achieves the highest overall success rate and consistently outperforms supervised, RL, and inference-time lookahead baselines on most websites.
Best results are in bold, and second-best are underlined.
}
\label{tab:webvoyager-results}
\centering
\setlength{\extrarowheight}{8pt}
\resizebox{\linewidth}{!}{%
\begin{tabular}{
@{}
p{4.2cm}
*{13}{>{\centering\arraybackslash}p{1.3cm}}
@{}
}
\multicolumn{14}{>{\columncolor{gray!15}}c}{
  \textbf{{\large \textit{Live Websites (WebVoyager)}}}
}
\\
\toprule
\textit{Method}
& \makecell[c]{\texttt{All}\\\texttt{Rec}}
& \makecell[c]{\texttt{Ama}}
& \makecell[c]{\texttt{Apple}}
& \makecell[c]{\texttt{ArXiv}}
& \makecell[c]{\texttt{BBC}\\\texttt{News}}
& \makecell[c]{\texttt{Camb}\\\texttt{Dict}}
& \makecell[c]{\texttt{Cour}\\\texttt{sera}}
& \makecell[c]{\texttt{ESPN}}
& \makecell[c]{\texttt{Git}\\\texttt{Hub}}
& \makecell[c]{\texttt{G.}\\\texttt{Map}}
& \makecell[c]{\texttt{GSea}\\\texttt{rch}}
& \makecell[c]{\texttt{HF}}
& \makecell[c]{\texttt{Wfram}\\\texttt{$\alpha$}} \\
\midrule

Llama3.1-8B-Instruct+CoT
& 2.4 & 2.5 & 0.0 & 9.5 & 7.6 & 11.8 & 0.0 & 0.0 & 1.2 & 4.3 & 0.0 & 2.1 & 5.2 \\

GPT-4o
& 31.1 & 25.8 & 29.7 & 27.9 & \underline{32.6} & 41.9 & 35.3 & \textbf{27.3} & \textbf{26.4} & 33.5 & \underline{5.3} & 19.4 & 37.1 \\

NNetNav
& 28.9 & 24.5 & 25.2 & 40.1 & 32.2 & 56.8 & 35.8 & 15.6 & 14.1 & 44.4 & 0.0 & 12.9 & 39.4 \\

Go-Browse
& 30.4 & 24.1 & 26.1 & 41.8 & 31.0 & 54.9 & 35.9 & 17.6 & 16.2 & 42.1 & 1.8 & 18.7 & 41.2 \\

WebRL
& \underline{32.6} & 20.9 & 27.3 & \textbf{50.4} & 29.7 & \underline{57.5} & \underline{36.6} & \underline{19.8} & \underline{18.4} & 40.6 & 2.7 & \underline{20.1} & \underline{43.7} \\

ITL
& 25.6 & \underline{28.7} & \underline{30.6} & 43.7 & 25.1 & 39.0 & 35.8 & 15.4 & 15.0 & \underline{46.5} & 0.0 & 1.6 & 18.8 \\

\midrule
\textbf{DynaWeb-8B (Ours)}
& \textbf{38.7} & \textbf{43.8} & \textbf{34.2} & \underline{45.7} & \textbf{39.8} & \textbf{61.6} & \textbf{38.3} & \underline{19.8} & 11.2 & \textbf{49.5} & \textbf{9.2} & \textbf{22.6} & \textbf{45.4} \\
\bottomrule
\end{tabular}%
}
\end{table*}

%% file: tex/5_analysis.tex
\section{Analysis}
We further analyze DynaWeb to better understand how key design choices in model-based dreaming affect training stability and overall agent performance.

\subsection{WM Dream Length Matters}
As discussed in Section~3, the length of synthetic trajectories generated through agent–world model interaction (referred to as the \emph{dream length}) plays a central role in determining learning effectiveness. From the perspective of sequential decision making, dream length controls the effective rollout depth, i.e., the number of imagined interaction steps over which the agent can accumulate credit and propagate reward signals. This induces a fundamental trade-off: trajectories that are too short limit the agent’s ability to learn multi-step action sequences required for non-trivial web tasks, while excessively long rollouts amplify compounding world model errors and hallucinations.

\begin{figure}[t]
  \centering
  \begin{subfigure}{0.48\linewidth}
    \centering
    \includegraphics[width=\linewidth]{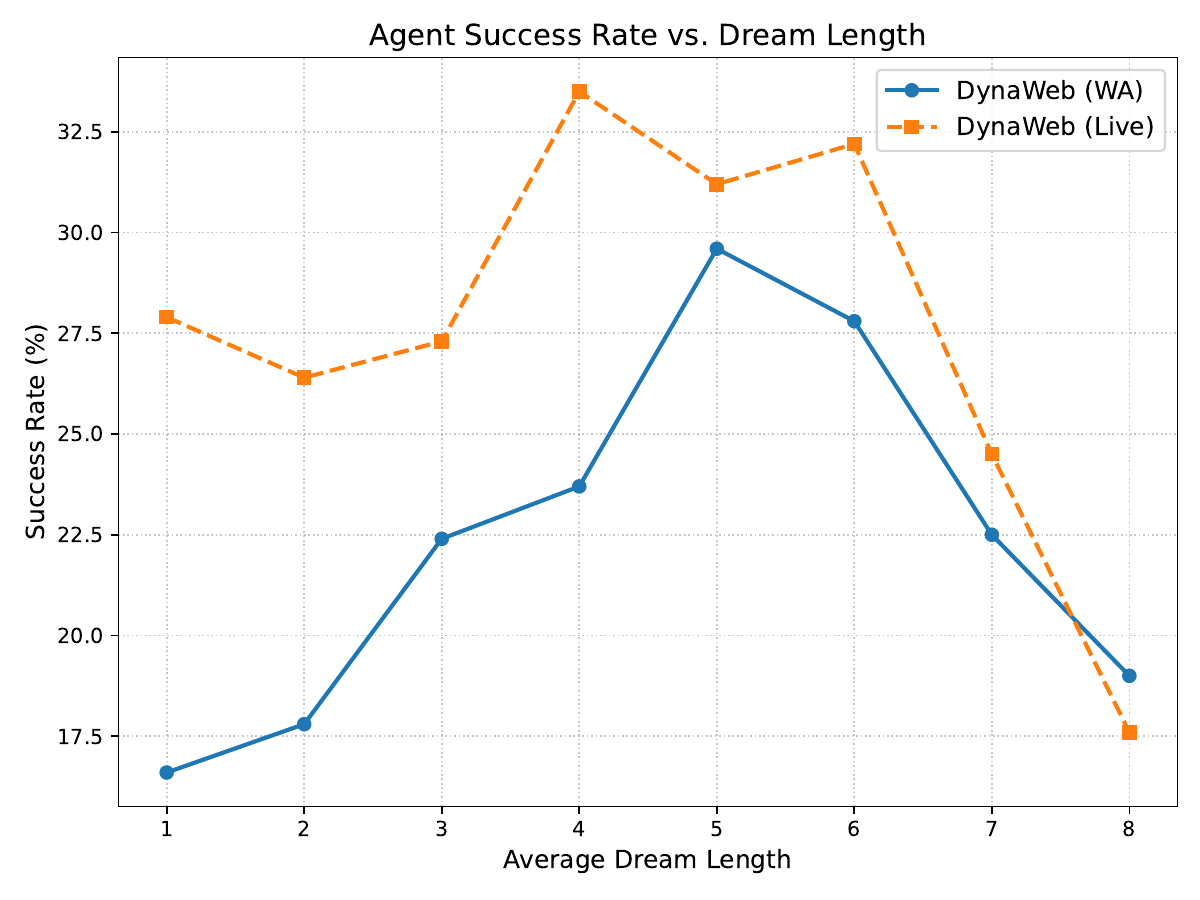}
    \caption{Agent success rate vs. dream length. A moderate length (4–5) performs best; shorter trajectories under-complete, while longer ones compound hallucinations.}
    \label{fig:dream_length}
  \end{subfigure}
  \hfill
  \begin{subfigure}{0.48\linewidth}
    \centering
    \includegraphics[width=\linewidth]{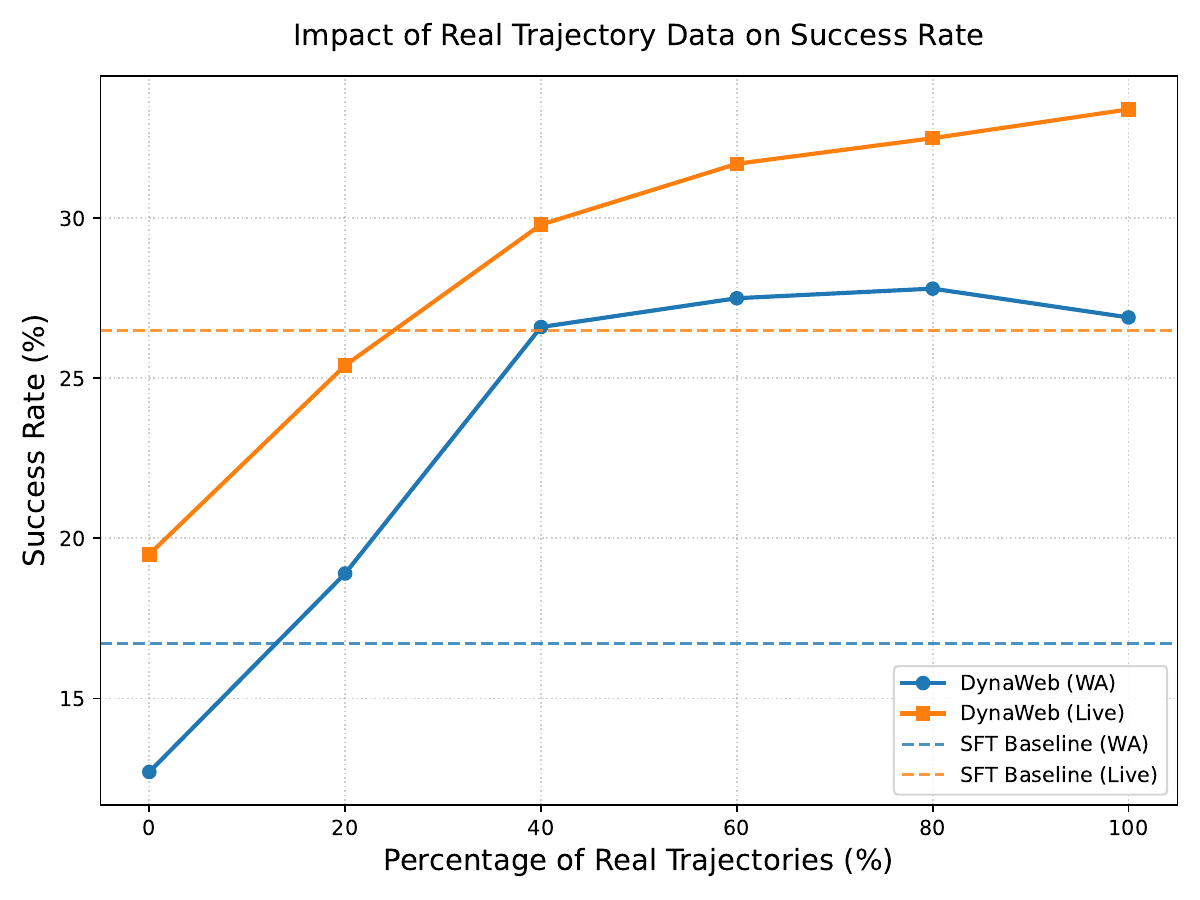}
    \caption{Impact of real trajectory data on success rate. Adding 40\% ground-truth real data significantly outperforms the SFT baseline, with diminishing returns for higher percentages.}
    \label{fig:real_data}
  \end{subfigure}
\end{figure}

To characterize this trade-off, we perform an ablation study in which DynaWeb agents are trained with varying maximum dream lengths. As shown in Figure\ref{fig:dream_length}, we observe a clear performance peak when the average dream length is around 4–5 steps. Shorter rollouts fail to capture sufficient interaction depth for task completion, whereas longer rollouts suffer from degraded training signals due to accumulated model inaccuracies. These results validate our design choice of restricting dream depth during training, and highlight that effective model-based web RL requires balancing rollout depth against simulation fidelity. Improving the robustness of world models to support longer-horizon, high-fidelity imagined trajectories remains an important direction for future work.

\subsection{Error Analysis of Local vs. Full-Page Transition Prediction}

Web state transitions exhibit heterogeneous difficulty. While many actions induce localized updates on the current page, others lead to full page transitions that substantially alter the entire accessibility tree. These two regimes differ significantly in terms of prediction complexity and error characteristics.

To better characterize this distinction, we partition the evaluation set into two subsets: (i) non-full-page-transition actions, where the resulting state can be described as a localized modification of the current page, and (ii) full page transition actions, which involve navigation to a substantially different page state.

For non-full-page-transition actions, we evaluate prediction quality using a sequence-level match rate against the ground-truth state-change description. Under this metric, the world model achieves a match rate of 97\%, indicating that it is highly reliable in modeling local updates, which constitute the majority of interaction steps.

We then analyze the more challenging subset of full page transition actions, which account for approximately 36\% of the dataset. In this regime, exact matching becomes less informative due to the large structural differences between states. Instead, we adopt an LLM-as-a-judge protocol using GPT-5 to assess prediction quality along multiple dimensions:

{\small
\begin{itemize}
    \item \textit{Action semantics}: correctness of understanding the intended effect of the action;
    \item \textit{Causal reasoning}: validity and consistency of the inferred transition logic;
    \item \textit{Resulting page correctness}: alignment of the predicted page with the ground-truth state;
    \item \textit{Faithfulness}: absence of hallucinated elements or content that should not appear.
\end{itemize}
}

Each dimension is scored in $[0,1]$, and the overall score is obtained by summing across the four criteria, yielding a final score in $[0,4]$. Under this protocol, the world model achieves an average score of 3.4/4 on full page transition actions.

Further inspection shows that, although discrepancies with the ground-truth accessibility tree occasionally arise, only about 12\% of cases are judged as completely incorrect, i.e., predictions that fail on multiple core dimensions and result in substantially wrong page states. The remaining cases typically preserve the high-level structure and semantics of the transition, despite minor inconsistencies at the element level.

Overall, these results indicate that the world model maintains high fidelity on local transitions and remains reasonably robust under full page transitions, despite the increased difficulty of the latter. This gap in difficulty also helps explain the sensitivity to rollout length observed in Section~5.1, as errors in full page transitions are more likely to propagate and compound over multi-step imagination.

\subsection{Effect of Real Interaction Data Percentage}
We analyze the role of real-environment interaction data in stabilizing and regularizing model-based web reinforcement learning. While DynaWeb relies heavily on imagined rollouts generated by the web world model, incorporating a portion of ground-truth trajectories from real environments can anchor learning to reliable state transitions and reward signals.

Figure~\ref{fig:real_data} reports the average success rate of DynaWeb agents under varying proportions of real interaction data. Two clear trends emerge. First, agents trained exclusively on simulated trajectories tend to underperform the SFT baseline, indicating that world model hallucinations and compounding errors can degrade learning when unregularized. Second, introducing approximately 40\% real trajectories leads to a substantial performance improvement over SFT, after which additional real data yields diminishing returns. 

\begin{wrapfigure}{l}{0.5\columnwidth}
  \vspace{-1.0em}
  \centering
  \caption{
  Effect of world model training on downstream agent performance.
  Success rate (\%) comparing a supervised task-specific world model (DynaWeb WM)
  and a frozen general-purpose LLM (GPT-oss-120b).
  }
  \vspace{-0.6em}
  \setlength{\extrarowheight}{3pt}
  \begin{tabular}{@{}lcc@{}}
    \toprule
    Benchmark & DynaWeb WM & GPT-oss-120b \\
    \midrule
    WebArena (Sim.) & \textbf{31.0} & 20.9 \\
    WebVoyager (Live) & \textbf{35.4} & 28.6 \\
    \bottomrule
  \end{tabular}
  \vspace{-0.8em}
  \label{tab:wm-training-ablation}
\end{wrapfigure}

These results suggest that real interaction data serves as a critical regularizer for imagination-driven training: a moderate amount is sufficient to stabilize learning and correct systematic model biases, while the majority of training signal can still be efficiently obtained from simulated rollouts. This balance highlights the practical advantage of DynaWeb in reducing reliance on costly real-environment interaction without sacrificing performance.

\subsection{Essential Role of WM Training}
We assess whether explicitly training a web world model is necessary for effective model-based reinforcement learning. To isolate the contribution of world model training, we keep the agent architecture, RL algorithm, and training protocol fixed, and replace the learned DynaWeb world model with a frozen GPT-oss-120b that is prompted to predict next-step observations during imagined rollouts.

As shown in Table~\ref{tab:wm-training-ablation}, agents trained with the supervised world model achieve substantially higher downstream success rates on both WebArena and WebVoyager. The large performance gap indicates that strong general-purpose LLM priors alone are insufficient to function as a reliable simulator for imagination-driven RL in web environments without learning environment-specific transition dynamics. Crucially, this result suggests that the effectiveness of DynaWeb does not arise merely from increased model capacity or prompting, but from grounding policy optimization in a world model explicitly trained to capture web interaction dynamics.

%% file: tex/6_conclusion.tex
\section{Conclusion}

We introduce \textbf{DynaWeb}, a model-based reinforcement learning framework that trains web agents through imagination rather than live web interaction.
By learning a web world model and using it to generate policy-driven imagined rollouts mixed up with real expert trajectories, DynaWeb enables effective on-policy optimization while avoiding the cost and risk of real-environment interaction.
Across WebArena and WebVoyager, DynaWeb consistently improves strong open-source web agents.

Our analysis reveals key principles for imagination-driven training, including the importance of rollout length, the regularizing role of real expert data, and the necessity of explicitly trained world models.
These results suggest that model-based imagination offers a practical and scalable path for training web agents, and point toward world-model-centric learning as a promising direction for long-horizon decision making in complex environments.

%% file: tex/appendix.tex
\section{More Details About DynaWeb}

\subsection{WebArena Training Prompt}

We provide the full system prompt used to train our web agent on WebArena.
The prompt defines the agent’s role, available actions, observation format, and task completion criteria, and is used consistently across both real-environment interaction and imagined rollouts generated by the web world model.
This prompt is fixed throughout training and evaluation to ensure reproducibility.

\begin{figure}[h]
\centering
\begin{tcolorbox}[width=\linewidth,
  colback=gray!5!white,
  colframe=gray!15!black,
  coltitle=black,
  colbacktitle=gray!25!white,
  fonttitle=\bfseries,
  fontupper=\footnotesize,
  title=WebArena Agent System Prompt
]
You are an AI assistant performing tasks on a web browser. You will be provided with task objective, current step, web page observations, interaction history and previous taked notes. You need to issue an action for this step.\\

Generate the response in the following format:\\
INTERACTION HISTORY SUMMARY:\\
Emphasize all important details in the INTERACTION HISTORY section.\\

OBSERVATION DESCRIPTION:\\
Describe information in the CURRENT OBSERVATION section. Emphasize elements and features that are relevant or potentially helpful for fulfilling the objective in detail.\\

REASON:\\
Provide your rationale for proposing the subsequent action commands here.\\

ACTION:\\
Select your action here.\\

OBSERVATION HIGHLIGHT:\\
List the numerical ids of elements on the current webpage based on which you would issue your action. Also include elements on the current webpage you would attend to if you fail in the future and have to restore to this step. Don't include elements from the previous pages. Select elements at a higher hierarchical level if most their children nodes are considered crucial. Sort by relevance and potential values from high to low, and separate the ids with commas. E.g., \texttt{1321, 52, 756, 838}.\\

You are ONLY allowed to use the following action commands. Strictly adheres to the given format. Only issue one single action.\\
Use the following actions:\\
- \texttt{click [id]}: To click on an element with its numerical ID on the webpage. E.g., \texttt{click [7]} If clicking on a specific element doesn't trigger the transition to your desired web state, this is due to the element's lack of interactivity or GUI visibility. In such cases, move on to interact with OTHER similar or relevant elements INSTEAD.\\
- \texttt{type [id] [content] [press\_enter\_after=0|1]}: To type content into a field with a specific ID. By default, the ``Enter'' key is pressed after typing unless \texttt{press\_enter\_after} is set to 0. E.g., \texttt{type [15] [Carnegie Mellon University] [1]} If you can't find what you're looking for on your first attempt, consider refining your search keywords by breaking them down or trying related terms.\\
- \texttt{stop [answer]}: To stop interaction and return response. Present your answer within the brackets. If and only if the task doesn't require a textual answer or appears insurmountable, indicate ``N/A'' and additional reasons and all relevant information you gather as the answer. Otherwise, including ``N/A'' will be penalized. E.g., \texttt{stop [5h 47min]}\\
- \texttt{go\_back}: To return to the previously viewed page.
\end{tcolorbox}
\caption{System prompt used for training and evaluation of the WebArena agent.}
\label{fig:webarena_prompt}
\end{figure}

\subsection{World Model System Prompt}

We provide the full system prompt used by the web world model.
The prompt specifies the input information available to the model, including the user objective, current webpage state, and executed actions, as well as the required output format for predicting web state changes and the resulting next-step accessibility tree.
This prompt is fixed throughout training and inference, and is shared across all experiments to ensure reproducibility.

\begin{figure}[htbp]
\centering
\begin{tcolorbox}[width=\linewidth,
  colback=gray!5!white,
  colframe=gray!15!black,
  coltitle=black,
  colbacktitle=gray!25!white,
  fonttitle=\bfseries,
  fontupper=\scriptsize,
  title=Web World Model System Prompt
]
You are an intelligent agent that predicts next state from given current action in a web environment, with your own logical reasoning.\\

Here's the information you'll have:\\
The user's objective: This is the task you're trying to complete.\\
The current web page's accessibility tree: This is a simplified representation of the webpage, providing key information.\\
The current web page's URL: This is the page you're currently navigating.\\
The previous action: This is the action you just performed in the previous step. It may be helpful to track your progress.\\
The current action: This is the current action that you performed to achieve the user's objective in the current web page's accessibility tree.\\
The format of previous actions can fall into several categories:\\

Page Operation Actions:\\
\texttt{click [id]}: This action clicks on an element with a specific id on the webpage.\\
\texttt{type [id] [content]}: Use this to type the content into the field with id. By default, the `Enter' key is pressed after typing unless press\_enter\_after is set to 0, i.e., \texttt{type [id] [content] [0]}.\\
\texttt{hover [id]}: Hover over an element with id.\\
\texttt{press [key\_comb]}: Simulates the pressing of a key combination on the keyboard (e.g., Ctrl+v).\\
\texttt{scroll [down]} or \texttt{scroll [up]}: Scroll the page up or down.\\

Tab Management Actions:\\
\texttt{new\_tab}: Open a new, empty browser tab.\\
\texttt{tab\_focus [tab\_index]}: Switch the browser's focus to a specific tab using its index.\\
\texttt{close\_tab}: Close the currently active tab.\\

URL Navigation Actions:\\
\texttt{goto [url]}: Navigate to a specific URL.\\
\texttt{go\_back}: Navigate to the previously viewed page.\\
\texttt{go\_forward}: Navigate to the next page (if a previous `go\_back' action was performed)\\

Completion Action:\\
\texttt{stop [answer]}: Done when you believe the task is complete.\\

Given the information above, you should first perform reasoning to predict expected changes on the current web page's accessibility tree,\\
and then generate the resulting next web page's accessibility tree based on your predicted web page changes.\\
Generate your answer in the following format:\\
\texttt{[Web state changes]}\\
changes\\
\texttt{[Next page accessibility tree]}\\
next\_acc\_tree\\

where \texttt{changes} are the predicted web page changes, and \texttt{next\_acc\_tree} is your predicted next page accessibility tree.\\
For next\_acc\_tree, you MUST generate a valid accessibility tree based on your predicted changes, do NOT output summary descriptions of how the next\_acc\_tree will change.\\
If the full predicted next\_acc\_tree is too long, you should output a pruned tree by removing unimportant elements that the web agent will unlikely use in subsequent action steps.\\
Meanwhile, you should NOT omit key unchanged elements that the web agent might interact with in future time steps.\\
Note: Even if the web page does not change, you must still output the complete next\_acc\_tree !
\end{tcolorbox}
\caption{System prompt used to train the web world model for predicting next-step accessibility trees from actions and current observations.}
\label{fig:world_model_prompt}
\end{figure}

\newpage

\subsection{Training Recipe and Hyperparameters}

This subsection summarizes the training recipe used for DynaWeb agent optimization.
Our training follows an online reinforcement learning setup in a synthetic web environment powered by a learned world model.
The agent is initialized from a supervised web navigation model and optimized using Group Sequence Policy Optimization (GSPO) with sparse terminal rewards.
Imagined rollouts generated by the world model are interleaved with real expert trajectories to stabilize learning.

\begin{table}[htbp]
\centering
\small
\begin{tabular}{l l}
\toprule
\textbf{Category} & \textbf{Setting} \\
\midrule
Backbone policy model & LLaMA-3.1-8B-Instruct \\

Optimization algorithm & GSPO (sequence-level policy optimization) \\
Advantage estimator & GRPO-style estimator \\
Policy learning rate & $1\times10^{-6}$ \\
Entropy coefficient & 0 \\
KL regularization & Disabled \\

Training epochs & 10 \\
Train batch size & 4 \\
Validation batch size & 32 \\

Max prompt length & 32{,}000 tokens \\
Max response length & 16{,}000 tokens \\
Max tokens per GPU (PPO) & 48{,}000 \\

Rollout engine & vLLM (asynchronous) \\
Rollout samples per prompt ($n$) & 8 \\
Rollout temperature & 0.7 \\
Rollout top-$p$ & 0.9 \\
Validation top-$p$ & 0.8 \\
Validation top-$k$ & 20 \\

World model max steps & 10 \\
Agent max steps & 10 \\

World model max tokens & 8{,}192 \\
World model temperature & 0.7 \\
World model top-$p$ & 0.9 \\

Precision & bfloat16 \\
Gradient checkpointing & Enabled \\
FSDP parameter offloading & Enabled \\
Optimizer offloading & Enabled \\

GPUs & 8 $\times$ H100 (single node) \\
\bottomrule
\end{tabular}
\caption{Key hyperparameters for DynaWeb agent training.}
\label{tab:training_recipe}
\end{table}